# How to Evaluate Your Question Answering System Every Day … and Still Get Real Work Done

Eric J. Breck, John D. Burger, Lisa Ferro, Lynette Hirschman,
David House, Marc Light, Inderjeet Mani

The MITRE Corporation
202 Burlington Rd., Bedford, Massachusetts 01730-1420 USA
{ebreck, john, lferro, lynette, dhouse, light, imani}@mitre.org

**Abstract**

In this paper, we report on Qaviar, an experimental automated evaluation system for question answering applications. The goal of our research was to find an automatically calculated measure that correlates well with human judges' assessment of answer correctness in the context of question answering tasks. Qaviar judges the response by computing recall against the stemmed content words in the human-generated answer key. It counts the answer correct if it exceeds a given recall threshold. We determined that the answer correctness predicted by Qaviar agreed with the human 93% to 95% of the time. 41 question-answering systems were ranked by both Qaviar and human assessors, and these rankings correlated with a Kendall's Tau measure of 0.920, compared to a correlation of 0.956 between human assessors on the same data.

## 1. Introduction

It would be nice to know if the work you did yesterday improved the performance of your system. However, if you spend the day evaluating the system performance by hand, there is no time left to make tomorrow's system better than today's. Therefore, automatic system evaluation is crucial to a tight, efficient development cycle. In this paper, we report on an experimental automated evaluation system for question answering applications. The goal of our research was to find an automatically calculated measure that correlates well with human judges' assessment of answer correctness in the context of several question answering tasks.

### 1.1. Question Answering

What do question answering systems attempt to do? Users often have specific questions which they hope or believe a particular resource can answer. This resource could be a particular text document, a collection of documents, a collection of web pages, a knowledge base of information, a semi-structured database, etc. The problem, from the user's perspective, is finding the desired answer. Question answering systems take a question as input and automatically provide one or more ranked answers to that question, based on the set of material available to the system. Ideally, we want to measure the answers in terms of being correct, justifiable, and concise, although, as we will see, we currently only measure correctness (see section 5 for more discussion of this).

A variety of disciplines within computer science have approached this task using different technologies and resources – see (Hirschman *et al.*, 1999), (Hirschman, 1998), (Chaudhri & Fikes, *eds.*, 1999), (Voorhees & Harman, 1999), (Green, 1969), and (Waltz, 1978) for examples. One view is to see question answering as a *service* provided for a collection of documents, as done in the TREC-8 Question Answering Track; the question answering system allows the user to pinpoint information in this collection. We can also look at question answering as a *demonstration* of understanding, as is done, for example, when students (or systems) are tested using standardized reading comprehension exams. In this case, the student (or system) is presented with a single passage and asked to answer questions about the information in the passage. The evaluation is done by comparing the system's or person's answers to an answer key (Hirschman *et al.*, 1999).

### 1.2. Automatic Evaluation

Regardless of approach or resource, all developers of question answering systems have the same question: when I modify my system, how do I know if I have improved it? Ideally, one would like to modify and evaluate frequently, searching efficiently for optimal configurations and testing out different design decisions. Automated evaluation can be invaluable for accomplishing this.

For instance you might look at questions like *what is the name of x?* or *what color is y?* and decide to build a system which answers *what* questions with nouns. But when you evaluate your system on a large corpus of questions, you will quickly discover that questions like *what did z do?* or *what kind of w is v?* are answered not by nouns but by verbs or adjectives. Conducting such a large evaluation repeatably and cheaply is greatly facilitated by having an automatically calculable metric.

Specifying an evaluation precisely enough that it can be carried out automatically is difficult regardless of the task. Even for seemingly objective tasks, many reasonable evaluations come to mind. For example, evaluating syntactic parsers involves measuring the difference between two syntactic parse trees, where one of them is assumed to be correct. A number of metrics have been proposed and used: labeled tree rate, consistent brackets recall rate, and consistent brackets tree rate (Black *et al*, 1991; Goodman, 1996). Issues include whether or not node labels matter, and how partial credit should be assigned.

In the Text Retrieval Conference, one issue is how to evaluate relevance of retrieved documents automatically, without having humans judge all the thousands of produced documents. A technique called *answer pooling* is used to produce a pool of relevant documents sufficiently comprehensive to estimate relevance of the document lists produced by each system (Sparck Jones & van Rijsbergen, 1975; Harman, 1994), within some known margin of error.

In question answering, the task is no less difficult. The space of possible answers is any string of text, and there is not always an obvious canonical answer nor a clear way to compare potential responses to it. The rest of this paper describes a prototype automatic evaluation system (section 2), an evaluation of our system (section 3), related research (section 4) and our conclusions and possible future work (section 5).

## 2. Qaviar

We have developed an evaluation tool Qaviar that marks system responses as correct or incorrect by comparing them to a human-generated answer key. Qaviar judges the response by computing recall against the stemmed content words in the human-generated answer key. It counts the answer correct if it exceeds a given recall threshold. Consider the example below.

Question:
  Who coined the term El Niño?

Answer Key:
  Peruvian fishermen
  => {peru fisherman}

System Response:
  Fisherman: They called it El Niño
  ⇨ {fisherman call niño}

The overlap between the system response and the answer key is one word: "fisherman" and there are two words in the answer key. Thus the answer key word recall is 1/2. It is not clear that a recall of 2/3 is twice as good as a recall of 1/3, and for comparison with human judgments (see section 3), a binary judgment is preferable. Therefore, Qaviar judges a system response as correct if this recall is above a preset threshold and incorrect otherwise.

The motivation for using this metric is that we expect that a good answer will contain certain keywords, but the exact phrasing does not matter. Removing stop-words and stemming the content words that remain is a further attempt to compensate for a difference in phrasing between the author of the answer key and the response produced by the system. Using recall without precision is justified if the responses are of roughly constant length, which was true for the TREC-8 Question Answering track: the track specified that an answer might be up to 50 bytes or up to 250 bytes. When we try to evaluate responses of varying lengths, it will become important to take into account precision, which begins to evaluate the conciseness of a response.

## 3. Evaluation of Qaviar

### 3.1. The TREC Data

We have evaluated Qaviar's predictions of human assessor's judgments. More specifically, we evaluated how well it predicts the correctness judgment that a NIST assessor assigned to a system response generated by a TREC-8 question answering track system. We have determined that the answer correctness predicted by Qaviar agrees with the human 93% to 95% of the time.

For the TREC Question Answering track, there were 37,927 system responses that were judged by the NIST judges; 35,684 of these responses were unique answer strings. There were 198 questions, and each run could provide up to 5 ranked answers to each question. 25 sites participated, each submitting one or both of a 50-byte-limited run and a 250-byte-limited run, for a total of 41 submitted runs.

Given these results, we asked our chief annotator to construct an answer key for the 198 questions. Our annotator had not been involved in any system development; she constructed the answer key based on her own knowledge, external resources like the Internet, plus the TREC system responses, and the TREC corpus. Consider the example below.

Question:
  Which company created the internet browser Mosaic?

Answer Key:
  National Center for Supercomputing Applications; NCSA | Netscape Communications

The key has the following interpretation: alternative forms of the same answer are separated by semi-colons, as in "National Center for Supercomputing Applications; NCSA". Different answers are separated by a vertical bar: "Netscape Communications" is a different but correct answer. In the answer keys generated by our annotator, there were on average 1.4 answers per question, 1.25 forms per answer, and 2.2 content words per answer form. In order to judge the correctness of a system response, Qaviar used the answer form that gave the highest recall for that response.

### 3.2. Raw Results on the TREC Data

Qaviar produces the same judgment as the NIST judges between 93% and 95% of the time, depending on where Qaviar sets its recall threshold for calling an answer correct. This range in accuracy is small because the vast majority of correct and incorrect system responses have a recall of 1 and 0, respectively. Table 1 below presents system response frequencies broken down by human judgment and recall.

| Recall Threshold | Human judged as incorrect | | Human judged as correct | |
|---|---|---|---|---|
| | Count | % of incorrect | Count | % of correct |
| 0.00 | 29709 | 92.4% | 336 | 5.8% |
| 0.01 to 0.25 | 325 | 1.0% | 36 | 0.6% |
| 0.26 to 0.50 | 1399 | 4.4% | 747 | 13.0% |
| 0.51 to 0.75 | 173 | 0.5% | 109 | 1.9% |
| 0.76 to 0.99 | 5 | 0.0% | 61 | 1.1% |
| 1.00 | 548 | 1.7% | 4479 | 77.7% |
| TOTAL | 32159 | 100.0% | 5768 | 100.0% |

Table 1: Human Judgment Compared to Qaviar

Note that incorrect responses outweigh correct ones. In fact, a baseline of 78% accuracy can be achieved by simply judging all responses as incorrect, since the majority of system responses in this data set are incorrect. If the rejection threshold is set so that Qaviar accepts responses with a recall of greater than 25% as *correct*, Qaviar would then misclassify 6.6% of the *incorrect* answers as correct, and 6.4% of *correct* answers as incorrect.

### 3.3. An ROC curve of the TREC Data

As the threshold varies, Qaviar's performance changes. A higher threshold means fewer false alarms, but also fewer identifications of truly correct responses. A lower threshold has the reverse effect. One way of presenting this variation graphically is called an ROC curve.

An ROC curve (Receiver Operating Characteristic), plots the *hit rate* versus the *false alarm rate* (Egan, 1975; Green & Swets, 1966). In other words, the horizontal axis is the percentage of time that Qaviar judges a system correct given that the human assessor judges the system incorrect, and the vertical axis is the percentage of time that Qaviar judges a system correct given that the human assessor also judged it correct. With a threshold of 1, Qaviar would always judge systems incorrect, giving 0% false alarms, but also 0% hits. With a threshold of 0, Qaviar would always judge systems correct, giving 100% hits, but also 100% false alarms. The curve is created by varying Qaviar's threshold from 0 to 1. For example, with the threshold of 0.25 discussed above, Qaviar performs with a hit rate of 93.6% and a false alarm rate of 6.6%; this point is plotted on the graph.

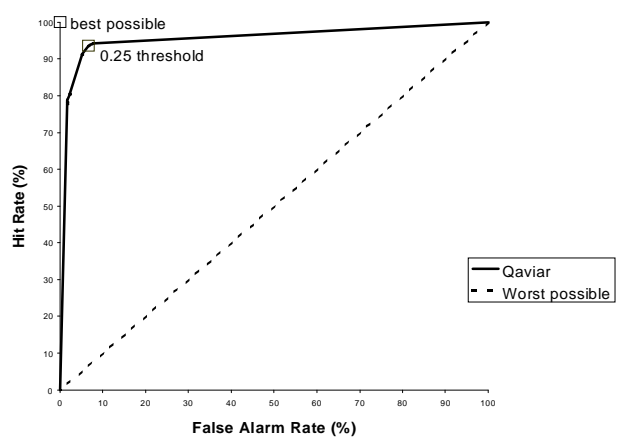

**Figure 2 - ROC Curve - Qaviar on TREC-8**

Another way of looking at an ROC curve is to look at the best possible and worst possible curves. The worst possible curve is a diagonal line, representing a "test" which randomly scores a response as correct *p*% of the time, where *p* is the varying parameter. The best possible test will always perform with a 100 % hit rate and a 0 % false alarm rate, so its curve is the point at the upper left side of the graph.

### 3.4. Correlation with TREC Rankings

Qaviar's scores of the 41 systems were used to rank them, and this ranking was compared to the official TREC ranking. The correlation between the two rankings, according to a metric called *Kendall's Tau*, was 0.920 (Stuart, 1983). We chose this metric because it was also used by the organizers of the TREC QA track to establish the stability of the results.

In order to establish that a single human judge could be used to judge the TREC-QA track, several stability correlations were performed by the organizers of the TREC QA task (Voorhees & Tice, 1999). In these experiments, the expensive, adjudicated three-judge-per-question rankings used officially in the 1999 TREC-QA evaluation were compared to (cheaper) single-judge-per-question rankings to see if the latter were close enough to substitute. The Kendall's Tau correlation metric was used to compare two rankings.

Kendall's Tau is calculated between two rankings *A* and *B*. Assuming that A ranks system x above system y, the pair (x,y) is called *concordant* if B also ranks x above y, and *discordant* if B ranks y above x. Kendall's Tau is the number of concordant pairs minus the number of discordant pairs divided by the total number of pairs, that is, for n = the number of systems:

$$\tau = \frac{concordant - discordant}{n(n-1)/2}$$

The average Kendall's Tau between a single-judge system and the official adjudicated three-judge system was 0.956. We compared a ranking of all 41 submitted runs (both 250-byte and 50-byte) produced by Qaviar to the ranking of the same systems by the official 3-judge judgments. The result of this Kendall's Tau correlation was 0.920.

### 3.5. Notes on this Evaluation of Qaviar

In this paper, we compare Qaviar to the human TREC assessors, but this comparison is slightly misleading. The TREC assessors' results are being used to compare the output of different systems, rank them, and choose a winner. As such, it is important that the assessors look carefully at each submitted system response to see if it might be judged correct. Qaviar, however, relies on there already being an answer key which it can use to score responses, and so it cannot adapt to novel responses. Since its answer key was constructed manually in part by looking at the answers judged correct by the TREC assessors, Qaviar's judgments are not independent of the assessors' judgments. In section 5 we suggest an experiment which does not have this difficulty.

On the other hand, there is no need for Qaviar to be independent from the TREC assessors, since it is not intended for year-end, multi-site evaluation. Instead, it is intended for use within one site during the development cycle, and as such it is appropriate to use the answer pools from the human assessors' judgments.

### 3.6. Failure Analysis

We performed a failure analysis of Qaviar using the TREC data set. A sample of 990 responses was randomly selected from the total 37,927 responses. We looked at the 72 responses where Qaviar and the assessor differed – cases where either the TREC assessor judged the answer correct but the recall was less than or equal to 0.5 or the TREC assessor judged the answer incorrect but the recall was greater than or equal to 0.5.

Results are shown in Table 3. Row (1) counts cases where it appeared to us that the TREC assessor's judgment was in error. Row (3) counts the examples where Qaviar was in error, which will be discussed in the next section. Row (2) counts the cases for which it was difficult to decide whether the TREC assessor or Qaviar was in error.

A significant fraction of Qaviar's errors came from disagreements between the TREC assessors and Qaviar on responses which contained information which was relevant to the question, but did not specifically answer it. Such responses contained all or part of the correct answer but additional extraneous information, so that a user might not be able to find the answer in the response. For example, to the TREC question:

*Who fired Maria Ybarra from her position in San Diego council?*

the following response was judged incorrect:

*… she was fired from her position as a council representative Friday in part because she became aware that Jim Sills , Henderson ' s top assistant…*

The correct answer was *Bruce Henderson*.

| (1) | TREC assessor (arguably) wrong | 7 | |
|---|---|---|---|
| (2) | Relevant responses ("tough call") | 27 | |
| (3) | Qaviar wrong: | 38 | |
| | Numeric expressions | | 19 |
| | Phrases | | 7 |
| | Other (stemming, stop-words, etc.) | | 12 |

Table 3: Error Analysis for Qaviar

### 3.7. Complexities of Automatic Scoring

The basic unit of recall as measured in Qaviar is the stemmed content word. While effective on the data so far, there are many cases where this unit is inappropriate and where comparison of units between a response and key is more complicated than string-matching.

#### 3.7.1. Numerical and Temporal Expressions

Phrases and digit strings which describe numbers and times may stretch across several words, and need to be compared differently than content words. The phrase *ten thirty* might be a single unit, and should compare as equal to either of two digit-expressions: the currency *$10.30* or the time *10:30*. *10%* should compare equal to *10 percent*, which should compare equal to *ten percent*. A relative date like *twenty years ago* needs to be resolved before it can be compared to an absolute date (e.g. *1980*) in an answer key. Finally, in many cases only a certain degree of precision is required, so even if an answer key were *1.4 billion*, a response of *1.39 billion* should be accepted.

#### 3.7.2. Phrases

In many cases, a person's surname is much more distinguishing than his or her given name. So while *Abraham* could apply to many politicians and would not be a good answer to *Who was the sixteenth president of the United States of America?*, the name *Lincoln* should probably suffice. Recall alone, however, would give either of these 0.50 if the answer key were *Abraham Lincoln*.

#### 3.7.3. Granularity

A question like *Where was George Washington born?* could be correctly answered by any of *Virginia*, *the United States,* or *Earth*, but only the first answer is particularly useful. Similarly, *When did George Washington die?* could be answered by *1799*, *the 18$^{th}$ century*, or *the second millennium*, but only the first (and possibly the second) are desirable answers. On the other hand, a different question (*When did George Washington live?*) might require a different degree of granularity.

#### 3.7.4. Context

A question like *Where is Rochester?* is not well-defined unless you know whether you're talking about Rochester, Michigan, Minnesota, New York or another of the over 20 Rochesters in the United States (let alone Rochesters in the rest of the world). *Who was the governor of Michigan?* is not well-defined unless you know what year is under discussion.

#### 3.7.5. Other Question Types

How will Qaviar work in scoring other question types? Consider the question

*Why did David Koresh ask the FBI for a typewriter?*

Then consider the following fragments of two responses which were scored correct.

*to enable him to record his revelations.*

*Koresh explained that he was delaying until he had finished writing the revelations of the Twelfth Seal.*

Given these two examples, it appears that the answer key should contain *revelations,* and then as alternates *write* and *record* as well. Even with both of those, a system might respond with yet a third synonym (e.g. *chronicle, transcribe, …*), for which it would not get credit.

When we consider more open-ended questions, like *What are the arguments for and against gun control?*, or *Who was Alfred Dreyfus?*, it remains an open question whether list-oriented answer keys (e.g., lists of topics) will suffice.

#### 3.7.6. Case and Stop-words

Even techniques as simple as case-folding and stopping can be problematic in key-response comparison. The word *in* is certain to appear on almost any stop-word list, so will be removed from system responses by Qaviar. If case is folded prior to stop-word removal, however, *IN* will also be removed, and that is the postal code for Indiana and a valid answer to the TREC question *Where is South Bend?*.

#### 3.7.7. Logical form

Even if the words match precisely, two sentences do not necessarily express the same content. *Man bites dog* and *dog bites man* are quite different. Logical negation is also not captured by recall.

#### 3.7.8. Possible Solutions to the Preceding Problems

Some of these difficulties might be alleviated by creating a more structured answer key. If the key were annotated with more- and less-important words (e.g. *Abraham **Lincoln***), weight might be distributed accordingly. Numerical answers might include a range of acceptable results (*1.35 - 1.45 billion*). In addition, an answer key might specify a constraint which the answer should satisfy, e.g., it should be a person name, an expression of a particular type, etc. However, this could impose a much greater burden on the creator of the answer key.

Problems with granularity and context might be dealt with by imposing guidelines on appropriate questions for evaluation. Perhaps *Where was George Washington born?* would have to be more clearly specified to be a valid question in this sort of evaluation. On the other hand, users might very well ask such questions, and so it seems undesirable to rule them out. It may be possible to introduce various degrees of precision in the scorer (this has been done in "minimum extent" named entity spans in MUC scoring), For example, a scorer could convert between different units, or "relax" a requirement for units.

The correct answers to *How* and *Why* questions will vary even more than the answers to other types of questions. While the current "bag of words" approach to scoring often performs reasonably, more intelligent scoring remains an open research problem.

## 4. Related Work

Martin and Lankester at the University of Ottawa also investigated automatic evaluation of the TREC-QA responses (1999). Their goal was to place a lower bound on a system's score, that is, to never score something incorrect as correct. Each response being judged by their automatic evaluator was compared to the set of responses which were judged correct by the TREC assessors. They began with the heuristic that a proposed answer is correct if it fully contains a correct TREC answer. They then revised this heuristic to address a number of problems, including stripped punctuation and whitespace, SGML tags, and document context. They encountered a number of tricky cases which we would do well to take account of. For example, sometimes the case of words matters. To the question: *What country is the biggest producer of tungsten*, the TREC assessors judged *China* to be a correct answer, but not *china*. Martin and Lankester do not report an evaluation of their technique.

Researchers at ETS developed E-Rater, a complex system to automatically score essay questions given on standardized tests. This system is currently in use, grading the Graduate Management Assessment Test's Analytical Writing Assignment (GMAT's AWA) since 10 February 1999. The following description is from their 1998 paper at the NCME Symposium on Automated Scoring:

> E-Rater's evidentiary feature scoring methodology incorporates more than 60 features that might be viewed as evidence that an essay exhibits writing characteristics described in the GMAT scoring guide. These variables comprise three general classes of features: syntactic, rhetorical, and topical content features. (Burstein *et al.*, 1998)

The features were combined with weights derived via stepwise linear regression to produce a single score.

While the e-Rater, like Qaviar, scores free-response answers to questions, its features for syntactic and rhetorical structure, as well as for scoring essays containing several arguments, are unnecessary for scoring short answers. However, the weighting which they use for their topical content features might be useful for scoring question-answering systems.

## 5. Conclusion

Our results lead us to believe that Qaviar will be useful during our system development cycle for TREC-style question answering tasks. We would like to know how well Qaviar works on different question sets such as the questions in reading comprehension exams (Hirschman *et al.*, 1999). An error analysis of the TREC data showed that numeric expressions and dates need to be normalized in some way. In addition, the concept of word overlap needs to score multi-word sequences appropriately (e.g., "secretary of state").

### 5.1. Future Work

One way to evaluate Qaviar on responses with an answer key independent of the document source would be to use Trivial Pursuit™ questions. We could run a question-answering system such as MITRE's Qanda (Breck *et al.*, 1999) on the questions, and then run Qaviar on Qanda's responses using the provided answer key (the back of the card). Qanda's responses could then be scored by a human and those judgments used to evaluate Qaviar.

Another domain of questions and answers is reading comprehension exams. A system has been developed to answer the questions on children's reading comprehension exams (Hirschman *et al.*, 1999). Further work in this area will be reported at the workshop on reading comprehension at the Applied Natural Language Processing conference in 2000, and in the reading comprehension group of the 2000 Johns Hopkins language engineering workshops. We plan to evaluate Qaviar's performance in scoring the results of these systems.

In an attempt to address the issues described in section 3.7, we hope to build a more complex feature-based model for automatic evaluation of question-answering systems. The system could include such features as the type of question (e.g. *Who*, *When*, or *Where*), the part of speech of words in the answer key, and whether there is a match of bigrams or trigrams between the answer key and the system's response. Such features could be combined using a loglinear model to learn appropriate weights.

#### 5.1.1. Beyond Correctness

Outside of the TREC context, as answers vary in length, recall is no longer appropriate on its own. Precision could be used to measure the *conciseness* of a response.

A good answer should directly answer the question. To the question, *How old is Clinton?* a response of *Bill Clinton, 46, was elected....* provides the correct information but does not exactly *answer* the question. Precision may be helpful in measuring this, but it may be necessary to consider more carefully what the goals are for evaluating a question-answering system.

It would be desirable for responses beyond a simple phrase to be *coherent*, which recall simply does not measure. It is not clear how coherence would be measured. Since the TREC QA task was largely extraction-based, it is also not clear whether the assessors considered the coherence of system responses in their scoring, so another data set might have to be used to evaluate the measure of coherence.

If justifications were included in the answer key, a response could be evaluated on how well it was justified. Justifications would be useful so that a user could know how much trust to place in a given answer, without always having to go and read the source document.

### 5.2. Goals of Evaluation

There are at least three different goals one might have in evaluating a question-answering system: *comprehension*, *utility for users*, and *utility for development*. Evaluation for comprehension attempts to determine whether the question-answerer *understands* the answer to the question. For example, if asked *Is Vincent Price dead?*, a response of *Vincent Price passed away on October 25, 1993* may not be acceptable, because while the response is relevant, there is an inference required to know that if a person has *passed away*, they are *dead*. A film database having one field, *alive*, with a binary value, could not accept the text string above without further processing.

If one were evaluating the utility to users, however, the response of *Vincent Price passed away on October 25, 1993* would be quite acceptable, because a user could immediately figure out that Mr. Price is dead. This response is good because it is true, it is concise (does not contain any extraneous information) and it also provides a bit of justification. This last is important because a practical system will fail reasonably often, and so if the system just said *Yes*, then the user would not know whether it could trust the system. It is interesting to note that a concise, justified truth is similar to the Platonic claim that knowledge is a *justified true belief*. Indeed, a reasonable goal for a user-centric evaluation might be that the user should be given knowledge about the answer to the question.

We want Qaviar to support iterative development and that demands a slightly different notion of acceptability. In a pure sense, only humans can judge utility for users and comprehension, but during the development cycle, as we have discussed, this can be prohibitively expensive. Therefore, the goal for a Qaviar should be to approximate these other goals; thus Qaviar should not differ *significantly* from these goals. That is, while Qaviar may make mistakes, we would like it not to make mistakes which would lead our system development astray. If the set of questions is large enough, such differences will hopefully be unimportant.

## 6. References

Black, E., Abney, S., Flickenger, D., Gdaniec, C., Grishman, R., Harrison, P., Hindle, D., Ingria, R., Jelinek, F., Klavans, J., Liberman, M., Marcus, M., Roukous, S., Santorini, B., Strzalkowski, T. (1991). A Procedure for Quantitatively Comparing the Syntactic Coverage of English Grammars. In *Proceedings of the Fourth DARPA Speech and Natural Language Workshop*, pp. 306-311.

Breck, E. Burger, J. D. Ferro, L. House, D. Light, M., Mani, I. (1999). A sys called Qanda. In (Voorhees & Harman, 1999).

Burstein, J., Kukich, K., Wolff, S., Lu, C., and Chodorow, M. (1998). Computer Analysis of Essays. NCME Symposium on Automated Scoring, April 1998.

Chaudri, V. and Fikes, R. co-chairs. (1999). *Proceedings of the 1999 AAAI Fall Symposium on Question Answering Systems*. North Falmouth, MA: AAAI Technical Report FS-99-02.

Egan, J. P. (1975). Signal detection theory and ROC analysis. Academic Press.

Goodman, J. (1996). Parsing Algorithms and Metrics. In *Proceedings of the 34th Annual Meeting of the ACL*. Santa Cruz, CA, June 1996.

Green, C. (1969). The Application of Theorem Proving to Question-Answering Systems. Ph.D. Thesis. Stanford University.

Green, D. M. and Swets, J. A. (1966). Signal detection theory and Psychophysics. John Wiley and Sons, Inc.

Harman, D. (1994). Overview of the Third Text REtrieval Conference (TREC-3). In *Proceedings of the Third Text Retrieval Conference (TREC-3)*. Gaithersberg, Maryland: NIST Special Publication 500-226.

Hirschman, L. (1998). Reading Comprehension: A Grand Challenge for Language Understanding. In *Proceedings of the first International Conference on Language Resources and Evaluation*.

Hirschman, L., Light, M., Breck, E., & Burger, J.D. (1999). Deep Read: A Reading Comprehension System. In *Proceedings of the Thirty-seventh Annual Meeting of the Association for Computational Linguistics* (pp. 325-332).

Martin, J., and Lankester, C. (1999). Ask Me Tomorrow: The NRC and University of Ottawa Question Answering System. In (Voorhees & Harman, 1999).

Sparck Jones, K. & van Rijsbergen, C. (1975). Report on the need for and provision of an "ideal" information retrieval test collection. British Library Research and Development Report 5266, Computer Laboratory, University of Cambridge.

Stuart, A. (1983). Kendall's tau. In Samuel Kotz and Norman L. Johnson, (Eds.), *Encyclopedia of Statistical Sciences*, volume 4, pp. 367-369. John Wiley & Sons.

Voorhees, E. M. and Tice, D. (1999). The TREC-8 Question Answering Track Evaluation. In (Voorhees & Harman, 1999).

Voorhees, E. and Harman, D. (1999). *Proceedings of the The Eighth Text REtrieval Conference* (TREC-8). Gaithersburg, MD: NIST Special Publication.

Waltz, D. (1978). An English language question-answering system for a large relational database. *Communications of the ACM*, 21, July 1978, pp. 526-539.